\documentclass{article}

\usepackage[preprint]{neurips_2022}

\usepackage[utf8]{inputenc} % allow utf-8 input
\usepackage[T1]{fontenc}    % use 8-bit T1 fonts
\usepackage[citecolor=green,colorlinks=true,linkcolor=blue]{hyperref}       % hyperlinks
\usepackage{url}            % simple URL typesetting
\usepackage{booktabs}       % professional-quality tables
\usepackage{amsfonts}       % blackboard math symbols
\usepackage{nicefrac}       % compact symbols for 1/2, etc.
\usepackage{microtype}      % microtypography
\usepackage{xcolor}         % colors
\usepackage{graphicx}         
\usepackage{bm}
\usepackage{wrapfig} 
\usepackage[labelformat=simple]{subcaption}

\usepackage{cleveref}   %可以调用 \Cref{fig:xxx}或者 \ref{fig:xxx} 
\setcitestyle{numbers,square,comma} %不加arxiv會報錯

\title{CLCNet: Rethinking of Ensemble Modeling with Classification Confidence Network}

\author{%
 Yao-Ching Yu$^1$$^,$$^2$  ~~~~  Shi-Jinn Horng$^1$$^,$\thanks{\scriptsize corresponding author\newline \hspace*{1.6em} Work supported by MOST of Taiwan under contract numbers 111-2218-E-011-011-MBK and 111-2221-E-011-134-, and also by the “Center for Cyber-physical System Innovation” from The Featured Areas Research Center Program within the framework of the Higher Education Sprout Project by the Ministry of Education (MOE) in Taiwan.} \vspace{0.5 em}\\
  $^1$Department of CSIE, National Taiwan University of Science and Technology \\
  $^2$AI Lab, Trend Micro \\
  yyq20130519@gmail.com~~~horngsj@yahoo.com.tw \\
}

\begin{document}

\maketitle

\begin{abstract}
In this paper, we propose a Classification Confidence Network (\emph{CLCNet}) that can determine whether the classification model classifies input samples correctly. The proposed model can take a classification result in the form of vector in any dimension, and return a confidence score as output, which represents the probability of an instance being classified correctly. We can utilize CLCNet in a simple cascade structure system consisting of several SOTA (state-of-the-art) classification models, and our experiments show that the system can achieve the following advantages: 1. The system can customize the average computation requirement (FLOPs) per image while inference. 2. Under the same computation requirement, the performance of the system can exceed any model that has identical structure with the model in the system, but different in size. In fact, we consider our cascade structure system as a new type of ensemble modeling. Like general ensemble modeling, it can achieve higher performance than single classification model, yet our system requires much less computation than general ensemble modeling. We have uploaded our code to a github repository: \href{https://github.com/yaoching0/CLCNet-Rethinking-of-Ensemble-Modeling}{https://github.com/yaoching0/CLCNet-Rethinking-of-Ensemble-Modeling}.
\end{abstract}

\section{Introduction}

In deep learning, classification \cite{yuan2021volo,liu2022convnet,brock2021high,tan2021efficientnetv2} has always been a popular task. And many SOTA classification models have been proposed in different size variants with the same structure, such as EfficientNet-B0 to B7 \cite{tan2019efficientnet}. The larger the number following B, the greater number of parameters the model has, along with better performance. Based on this observation, we want to achieve the following purposes: 
\begin{itemize}

\item For a classification model with variants in different sizes, we expect to propose a method that can combine variants of the model, which can achieve higher accuracy than the original model with the same requirement of computation (FLOPs).
\item The above method can achieve the performance of general ensemble modeling, but the demand for computation is lower.
\end{itemize}

In order to achieve the above purposes, firstly, we propose a network that can predict whether the classification model classifies correctly, called Classification Confidence Network (\emph{CLCNet}). Given the output result of a classification model in the form of vector and pass it to CLCNet, CLCNet will return a confidence score representing the correctness of the classification result. The higher the score, the higher the probability that CLCNet thinks the classification is correct, and vice versa. In particular, in CLCNet, we adopt a special mapping mechanism, so that CLCNet can accept classification results of any dimension. When transferring CLCNet from one classification task to another classification task with a completely different number of categories, it can even be used immediately \emph{without} retraining.

% Then we can apply CLCNet in a simple cascade structure system, in which we stack two (or several) classification models as shown in Fig.\ref{fig1}. The model required less computation is called shallow model, and the model with higher computation cost is called deep model. When inferring a sample through this system, it will first use the shallow model to classify and return its classification result to CLCNet to predict whether it is classified correctly. If CLCNet outputs a high confidence score, we will accept the classification result directly and won’t perform subsequent steps. On the other hand, when CLCNet has insufficient confidence to the classification result of the shallow model, the input sample will proceed to be classified by the deep model, and the classification result of deep model will also be evaluated by CLCNet and output a confidence score. Finally, we compare the confidence scores predicted by CLCNet on classification results of both deep model and shallow model, and return the classification result with the higher confidence as output.

\begin{figure}
  \vspace{-0em}  
  \centering
  \includegraphics[width=13.5cm]{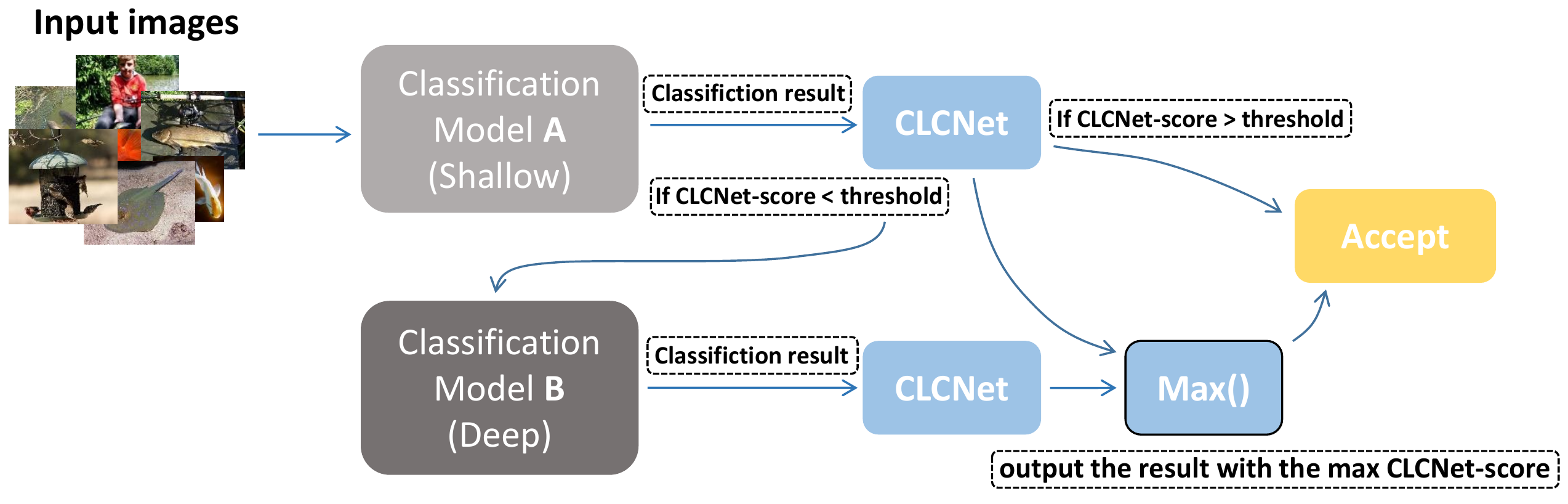}
  \caption{The illustration of cascade structure system.}
  \label{fig1}
  \vspace{-1em}  % 調整上下文間距
\end{figure}

Then we can apply CLCNet in a cascade structure system, in which we stack two classification models as shown in Fig.\ref{fig1}. This system let the simple input samples to be classified by the shallow (lightweight) model first, and only a small number of difficult samples with low confidence are further classified by the deep model. It can greatly save the amount of computation, and at the same time ensure that the accuracy is almost not reduced, or even better. Many studies \cite{wang2017idk,enomoro2021learning,wan2018confnet} have demonstrated the effectiveness of the cascaded structure system.

In fact, we can also directly use the highest probability of the classification result as the confidence score of the classification. For example, in the classification result of the five-class classification problem $(0.6, 0.1, 0.1, 0.1, 0.1)$, directly use 0.6 as the confidence score. However, it is not able to distinguish such a case. If there is another classification result $(0.6, 0.39, 0.01, 0.0, 0.0)$, it is obvious that the highest probability of these two classification results is 0.6, but the second highest probabilities are 0.1 and 0.39, respectively. Intuitively, the classification result with the second highest probability of 0.39 is more likely to be misclassified, so it should have a slightly lower confidence score. In order to be able to distinguish this situation and evaluate it better, we propose CLCNet, expecting it to make a prediction based on the numerical distribution of a classification result.

We will evaluate this cascade structure system on the dataset ImageNet \cite{deng2009imagenet}, and the results will show that for the single models (such as shallow or deep model) in the system, the system can outperform them with the same amount of computation. Or achieve the same accuracy, while the system required less computation than that of single models. We will also show that the system can also reach the performance of general ensemble modeling, that is, to obtain higher accuracy than any single model in the system, but with less computation than general ensemble modeling.

\section{Related work}
\label{Related_work}
ConfNet \cite{wan2018confnet} has tried to give a confidence for the classification result to represent whether the classification is correct. It directly adds a fully connected layer at the end of the classification model. When the classification model finishes classifying an input sample, the output of the classification model is sorted from largest to smallest to remove the category information, and then input the sorted classification result to the fully connected layer with sigmoid activation function, the network will eventually return a value ranged from 0 to 1, which represents the confidence of this classification.

ConfNet is an end-to-end network, it will output a classification result and corresponding confidence at the same time. During training the ConfNet, it will use the Confidence Loss \cite{wan2018confnet} and this loss will be calculated based on both classification result and confidence score. Its purpose is to constrain network to give a lower confidence when cross-entropy loss of a classification output is larger, and vice versa.

This end-to-end model has a distinct disadvantage. Because the classification model and the subsequent fully connected layer are trained at the same time, when doing gradient descent, the previous classification model must consider not only to get correct classification, but also when the cross-entropy loss is low, its classification output needs to "\emph{look like}" to be a correct classification for the subsequent fully connected layer to recognize. In contrast, when the cross-entropy loss is high, the classification model must make its classification result "\emph{look like}" a misclassification result, which may damage the performance of the classification model itself. Moreover, because the shape of the final fully connected layer is fixed and can only accept fixed-dimensional input, whenever the classification task changes, the entire classification model and the final fully connected layer need to be retrained.

The above problems incurred in ConfNet \cite{wan2018confnet} will not appear in CLCNet, CLCNet and the classification model are trained separately, so the performance of the classification model will not be damaged. And it adopts a special mapping mechanism, which can accept a classification result of arbitrary dimension as input. The training set of CLCNet can contain the outputs of multiple different classification models to enhance its robustness. After CLCNet is trained, it can be used directly without fine-tuning when transferring to new tasks in the future, and good performance can still be achieved. Of course, fine-tuning CLCNet on new datasets may achieve better performance.

\section{Proposed method}
\label{Proposed_method}
In this chapter, we first introduce network architecture and technical details of CLCNet. Then, we will describe the features of the cascade structure system.
\subsection{CLCNet (Classification Confidence Network)}
\begin{figure}
  \vspace{-0em}  
  \centering
  \includegraphics[width=13.5cm]{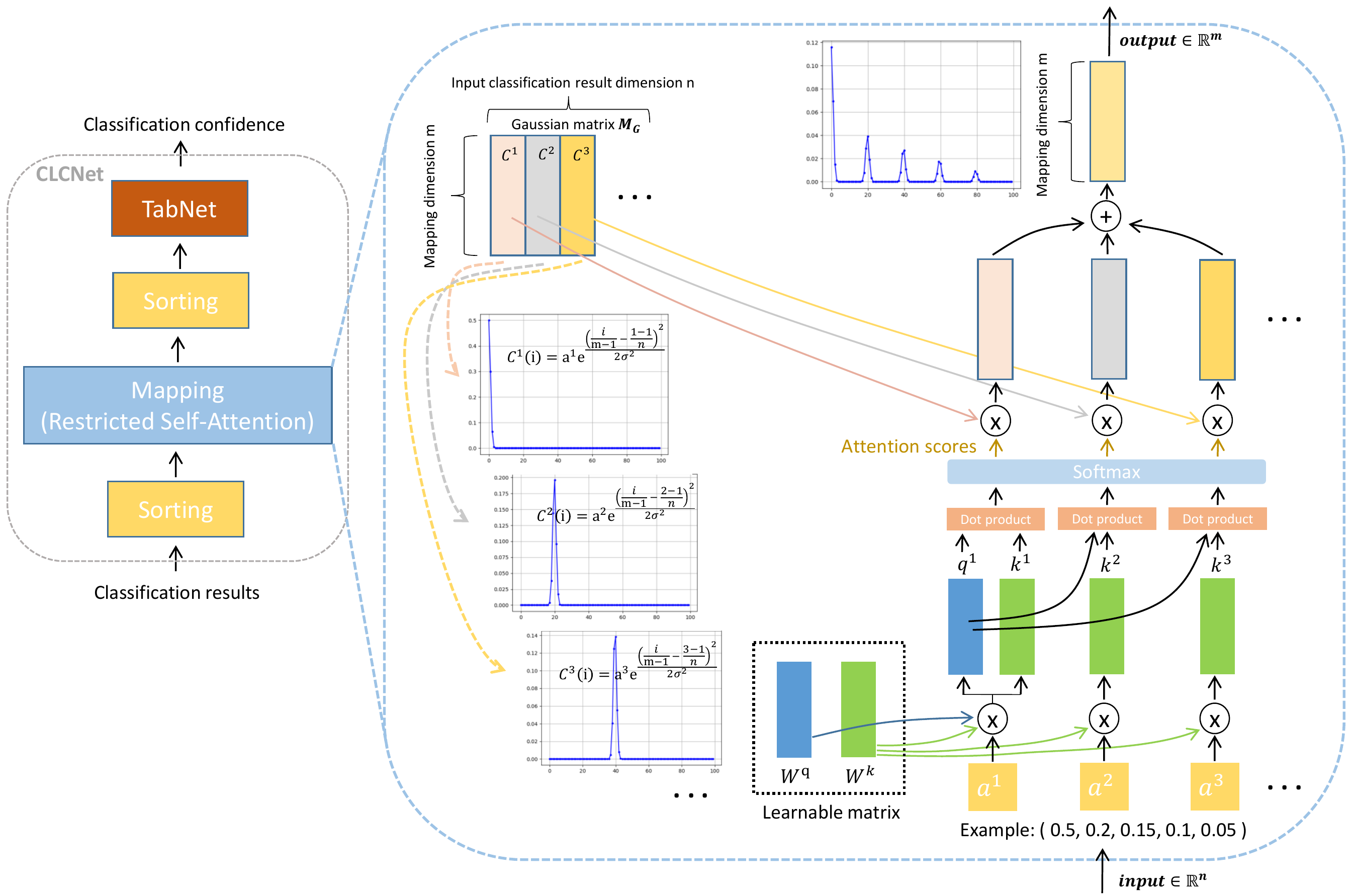}
  \caption{The architecture of CLCNet (left) and the schematic of Restricted Self-Attention (right).}
  \label{fig2}
  \vspace{-0em}  % 調整上下文間距
\end{figure}

When we get a classification result, we want to know whether the classification is correct. We will input the classification result to CLCNet which is shown on the left side of Fig.\ref{fig2}. First CLCNet will sort the class probability of the classification result from largest to smallest to remove its category information, in this way, it can focus on the numerical distribution of the classification result. Then proceed input the sorted result into proposed Restricted Self-Attention module, the purpose of which is to map a classification result of any dimension to an equivalent $m$-dimension vector, where $m$ is a hyperparameter. In other words, we want to "\emph{simulate}" an equivalent numerical distribution of the classification result in $m$-classification task. When the equivalent $m$-dimensional vector is obtained, we sort it again and give it to TabNet \cite{arik2021tabnet}, TabNet will return a confidence score for this mapping vector, representing the probability that the classification is correct.

\paragraph{Restricted Self-Attention}
The architecture of Restricted Self-Attention is shown on the right side of Fig.\ref{fig2}, its $\bm{input}\footnote{We denote vectors or matrices in bold.}\in\left\{\bigcup_{i=1}^{\infty} \mathbb{R}^{i}\right\}$, and its $\bm{output} \in \mathbb{R}^{m}$, $m$ is a hyperparameter, we set it to 100 by default. It is a variant of Self-Attention in Transformer \cite{vaswani2017attention}, so some parts of the inference process are the same. Given a sorted classification result input as $\bm{a} \in \mathbb{R}^{n}$, which is $(a^{1},a^{2}, \ldots ,a^{n})$, first we use a learnable matrix $\bm{W^{q}}\in\mathbb{R}^{m\times 1}$ to get $\bm{q^{1}} = \bm{W^{q}}\cdot a^{1}$. Next, multiply $a^{x}$ with another learnable matrix $\bm{W^{k}}\in\mathbb{R}^{m\times 1}$, where $x \in\{1,2,3, \ldots, n\}$, to get $\bm{k^{x}} = \bm{W^{k}}\cdot a^{x}$. Then we use Eq.\ref{eq_1} to obtain the attention score $\bm{att}\in\mathbb{R}^{n\times 1}$, which has the same length as the input $\bm{a}$ ($att^{x}$ is the attention score of $a^{x}$).
\begin{equation} \label{eq_1}
 \bm{att} = softmax\left(\left[\bm{k^{1}}, \bm{k^{2}}, \ldots, \bm{k^{n}}\right]^{T} \cdot \bm{q^{1}}\right)
\end{equation}

$[~\cdot,\cdot~]$ means to concatenate the vectors horizontally. Next, we define a rule-based matrix $\bm{M_{G}}\in\mathbb{R}^{m\times n}$. Obviously, the matrix consists of $n$ $(m\times 1)$-dimensional column vectors, and we use $\bm{C^{x}}\in\mathbb{R}^{m\times 1}$ to represent one of the column vectors of the matrix, so $\bm{M_{G}} = [\bm{C^{1}},\bm{C^{2}},\ldots,\bm{C^{n}}]$. Further, the value in $\bm{C^{x}}$ satisfies the following equation:
\begin{equation} \label{eq_2}
\bm{C^{x}}(i)=a^{x}\cdot e^{-\frac{\left(\frac{i}{m-1}-\frac{x-1}{n}\right)^{2}}{2 \sigma^{2}}}, i \in\{0,1,2, \ldots, m-1\}
\end{equation}
$\bm{C^{x}}$ is a vector composed of the function values of m points on Eq.\ref{eq_2}. In fact, the function (Eq.\ref{eq_2}) is a simplified version of the Gaussian probability density function, which is also a bell shaped function, i represents the position in the $\bm{C^{x}}$, $n$ is the dimension of the input classification result, $\frac{x-1}{n}$ is the position of the maximum value of the bell shaped function (i.e. mean value ($\mu$) of the distribution represented by the simplified probability density function), so the function corresponding to different $x$ will have different position of the maximum value, $\sigma$ is a hyperparameter that can control the shape of the function (i.e. the standard deviation of the distribution represented by the simplified probability density function), function with a larger $\sigma$ will have a flatter shape, and vice versa. We set $\sigma$ to 0.01 by default, and we draw examples of $\bm{C^{1}}$, $\bm{C^{2}}$, and $\bm{C^{3}}$ assuming $n$ being set to 5 in Fig.\ref{fig2}. With Eq.\ref{eq_2} we can obtain all $\bm{C^{x}}$  and concatenate them by column to form matrix $\bm{M_{G}}$ .

Now, we can use the already obtained $\bm{att}$ and $\bm{C^{x}}$ to calculate the mapping vector output:
\begin{equation} \label{eq_3}
 \bm{output} =\sum_{x=1}^{n} att^{x} \cdot \bm{C^{x}}
\end{equation}

Pragmatically we will also use matrix operations directly to speed up:
\begin{equation} \label{eq_4}
 \bm{output} =\bm{M_{G}} \cdot \bm{att}
\end{equation}

\begin{figure}
  \vspace{-0em}  
  \centering
  \includegraphics[width=13cm]{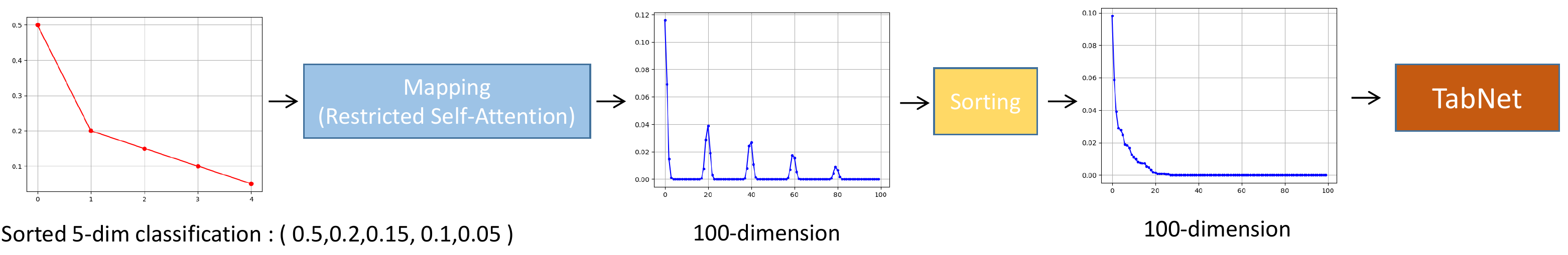}
  \caption{An example of mapping a $5$-dimensional classification result to an equivalent $100$-dimensional vector.}
  \label{fig3}
  \vspace{-1em}  % 調整上下文間距
\end{figure}

As mentioned at the beginning, $\bm{output} \in \mathbb{R}^{m}$, and we sort it again to get the final mapping vector. Finally, we show an example where $5$-dimensional classification result is mapped to a $100$-dimensional vector in Fig.\ref{fig3}.

\paragraph{How does this mapping work}
First of all, the role of the learnable matrices $\bm{W^{q}}$ and $\bm{W^{k}}$ is to give different $a^{x}$ attention scores to appropriately adjust their proportions in the final mapping vector. Specifically, although $\sum_{x=1}^{n} att^{x} = 1$ and $\sum_{x=1}^{n} a^{x} = 1$, $att^{x}$ does not need to be equal to $a^{x}$, and $a^{x}/a^{y}$ is not necessarily equal to $att^{x}/att^{y}$.

\begin{figure}
  \vspace{0em}  
  \centering
  \includegraphics[width=13.5cm]{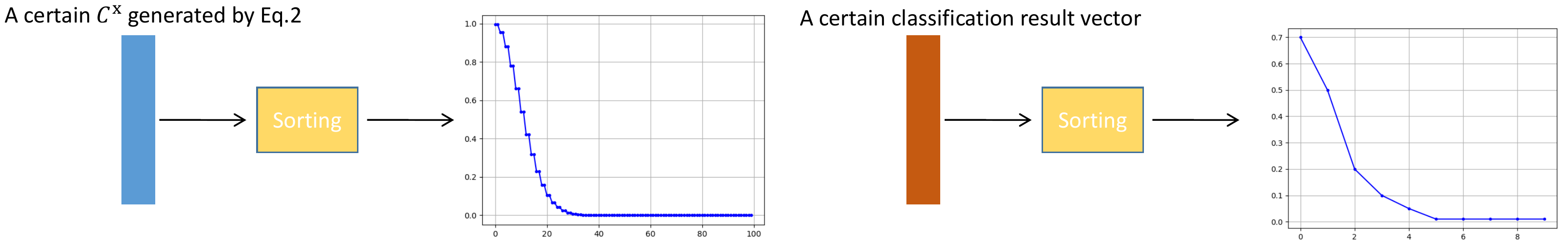}
  \caption{The shapes of a sorted $\bm{C^{x}}$ and a sorted classification vector.}
  \label{figy}
  \vspace{-1em}  % 調整上下文間距
\end{figure}

\begin{wrapfigure}{r}{0.52\textwidth}
  \vspace{-1em}  
  \centering
  \includegraphics[width=6.5cm]{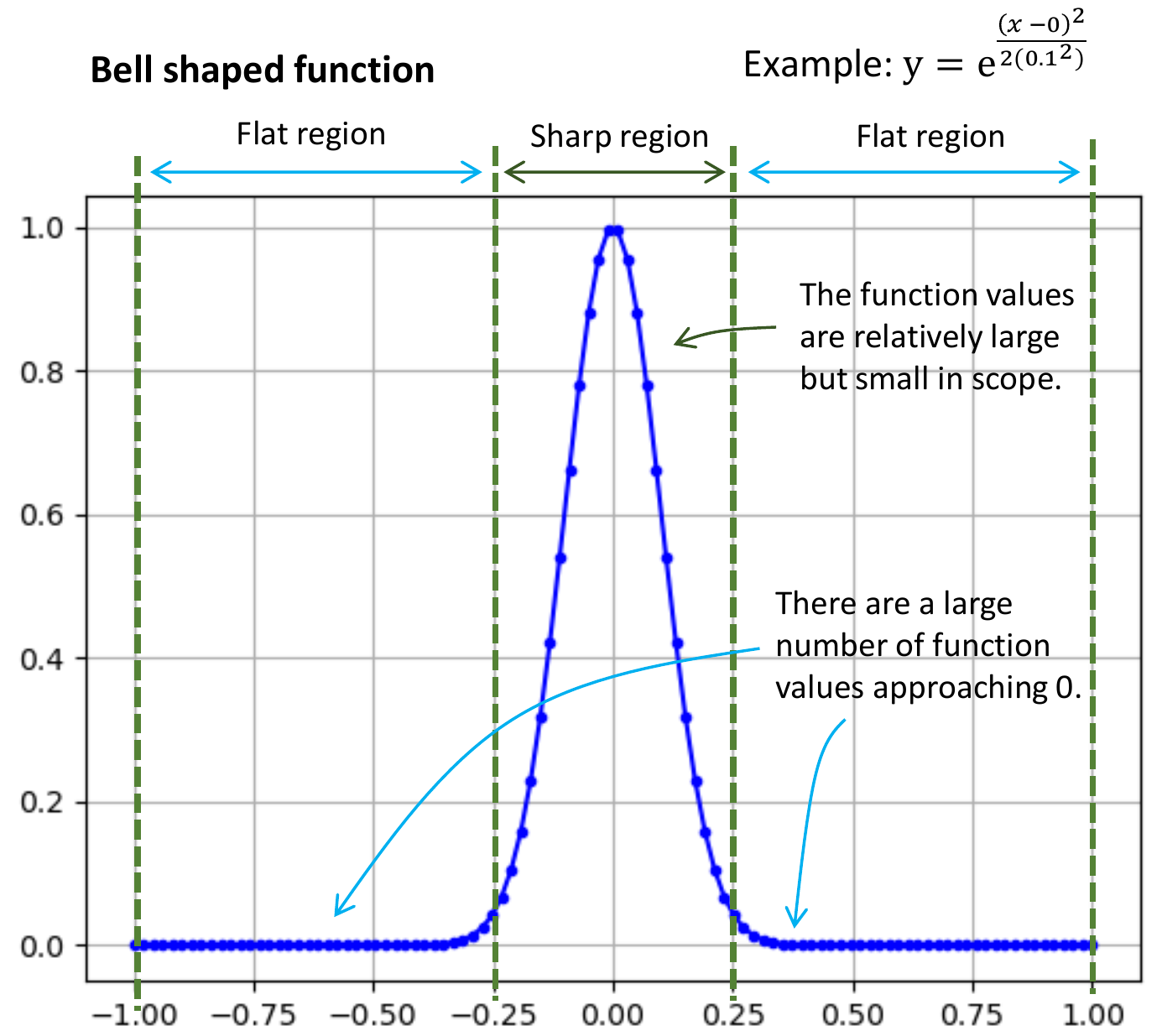}
  \caption{Schematic diagram of bell shaped function.}
  \label{figx}
  \vspace{-1em}  % 調整上下文間距
\end{wrapfigure}

We choose the bell shaped function as the function of $\bm{C^{x}}$ because its numerical distribution is very similar to that of a general classification result vector. A bell shaped function can be divided into two distinct parts, as shown in Fig.\ref{figx}, one is a sharp region close to the maximum ($\frac{x-1}{n}$ in $\bm{C^{x}}$), which is small in scope but large in function values. The other part is the flat areas on the left and right away from the maximum value, where there are a large number of function values approaching zero. And according to the observation, the sorted classification vectors in real-world cases are mostly in the form of $(0.8, 0.1, 0.00034, 0.00022, 0.00013, ...)$, that is, those vectors have a very small amounts of large probability values, and have a very large amounts of small probability values close to zero, which is very similar to the distribution of $\bm{C^{x}}$, and the large probability values correspond to the sharp region of $\bm{C^{x}}$, the small probability values correspond to the flat region of $\bm{C^{x}}$. So if we sort $\bm{C^{x}}$ from largest to smallest, it will be very similar to a sorted general classification vector, as shown in Fig.\ref{figy}, and because of this, we can use the sorted $\bm{C^{x}}$ to "\emph{simulate}" a sorted classification vector.

We can regard each $\bm{C^{x}}$ as a mapping vector of $a^{x}$ on the $m$-classification task, and $\bm{C^{x}}$ will be multiplied by $att^{x}$ (Eq.\ref{eq_3} and Fig.\ref{fig2}). The $att^{x}$ is generated by considering the relative size of $a^{x}$ in $\bm{a}$, and is responsible for adjusting the overall value of $\bm{C^{x}}$. The maximum value of $att^{x} \cdot \bm{C^{x}}$ will be $att^{x} \cdot a^{x}$ (according to Eq.\ref{eq_2}, the maximum value of $\bm{C^{x}}$ is $a^{x}$). Next, we add all $att^{x} \cdot \bm{C^{x}}$, it can be said that each $a^{x}$ participates in the generation of the final mapping vector. And because the positions of the maxima of different $\bm{C^{x}}$ are different (i.e. $\frac{x-1}{n}$), when we finally sort the $\sum_{x=1}^{n} att^{x} \cdot \bm{C^{x}}$, the sharp regions from each $att^{x} \cdot \bm{C^{x}}$ can be partially preserved (as shown in Fig.\ref{fig3}, the sharp areas are staggered from each other). In this way, the more large probability values in $\bm{a}$, the more and denser the sharp areas in $\sum_{x=1}^{n} att^{x} \cdot \bm{C^{x}}$, and the more large probability values will be in the final mapping vector. Finally, we sort the $\sum_{x=1}^{n} att^{x} \cdot \bm{C^{x}}$, as mentioned before, the sorted $\bm{C^{x}}$ is very similar to a sorted general classification vector, and so is the sorted $\sum_{x=1}^{n} att^{x} \cdot \bm{C^{x}}$.

\begin{figure}
  \vspace{0em}  
  \centering
  \includegraphics[width=13.5cm]{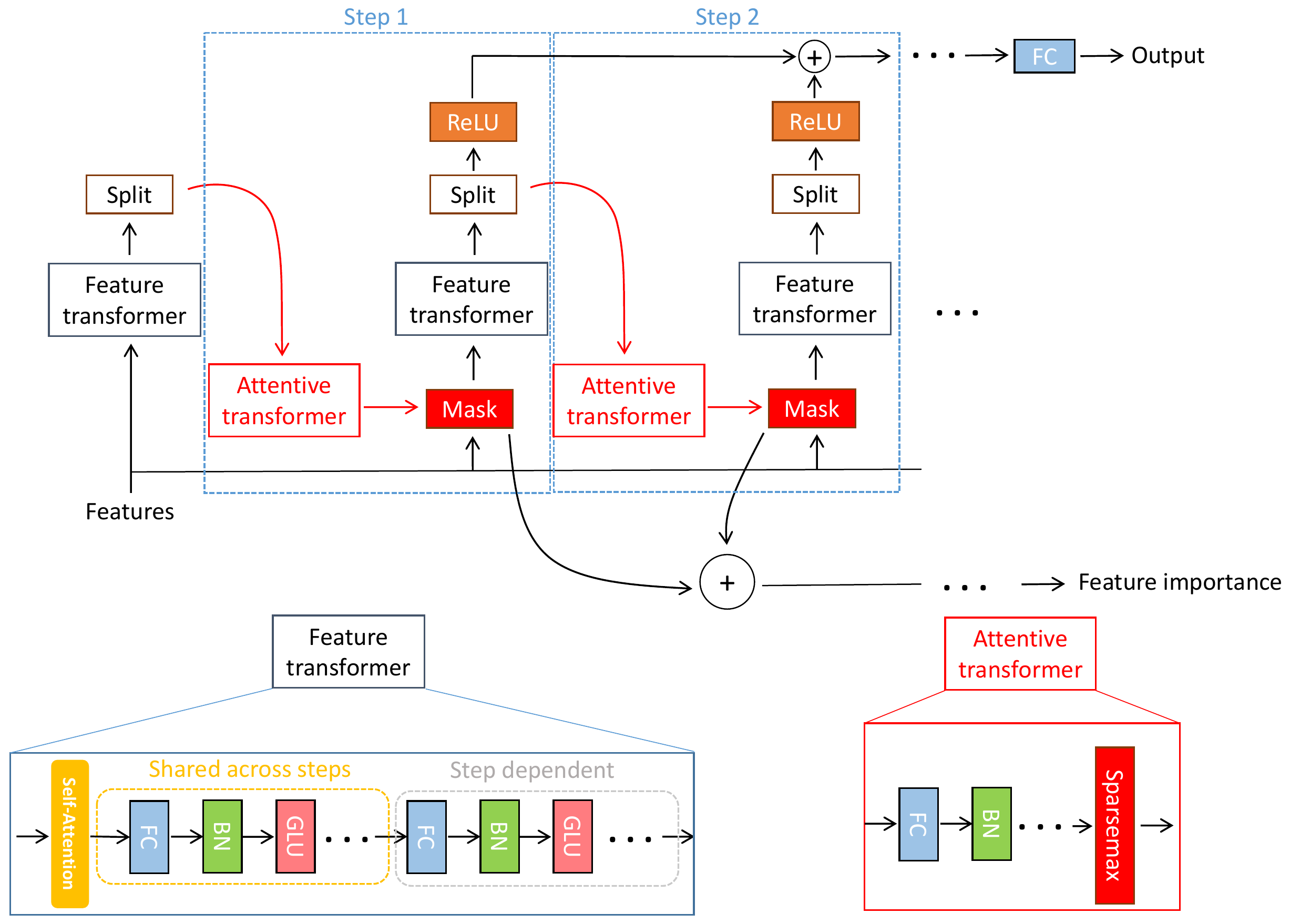}
  \caption{Architecture of TabNet.}
  \label{fig4}
  \vspace{-0em}  % 調整上下文間距
\end{figure}

\paragraph{TabNet}
After mapping and sorting, we get an $m$-dimensional vector, and we will continue to input it to TabNet \cite{arik2021tabnet} to get a confidence score. We can treat the $m$-dimensional vector as a piece of tabular data. TabNet can approximate the performance of tree-based models (such as LGBM \cite{ke2017lightgbm} and XGBOOST \cite{chen2015xgboost}) on regression task of tabular data without requiring additional feature engineering. The structure of TabNet is shown in Fig.\ref{fig4}. TabNet takes many steps which consist of the same structure while inference. Given a set of input features (an m-dimensional vector can be regarded as m features), when starting a step, a mask will be generated by the attentive transformer \cite{arik2021tabnet} (see Fig.\ref{fig4}). The mask indicates which features will be selected in this step. Next, continue to input the features selected by the mask into the feature transformer \cite{arik2021tabnet} (see Fig.\ref{fig4}), and its output will be split into two parts, the first part is responsible for determining the confidence score, and the second part is inputted to the attentive transformer to determine the mask shape of the next step. When all steps are executed, the first part of feature transformers’ output in each step is summed up, and after a fully connected layer, the final confidence score can be obtained.

It should be noted that we have modified the original TabNet by adding a Self-Attention \cite{vaswani2017attention} module shared across different steps before the first layer FC of the feature transformer, which can alleviate the instability caused by the sparsity of the input features of the feature transformer. The sparsity is cause by the last layer of the attentive transformer that determines the mask, which uses Sparsemax \cite{martins2016softmax} as activation function, therefore tends to output a sparse mask.

\newpage
\subsection{Cascade structure system}

\begin{wrapfigure}{r}{0.4\textwidth}
  \vspace{-2em}  
  \centering
  \includegraphics[width=5.5cm]{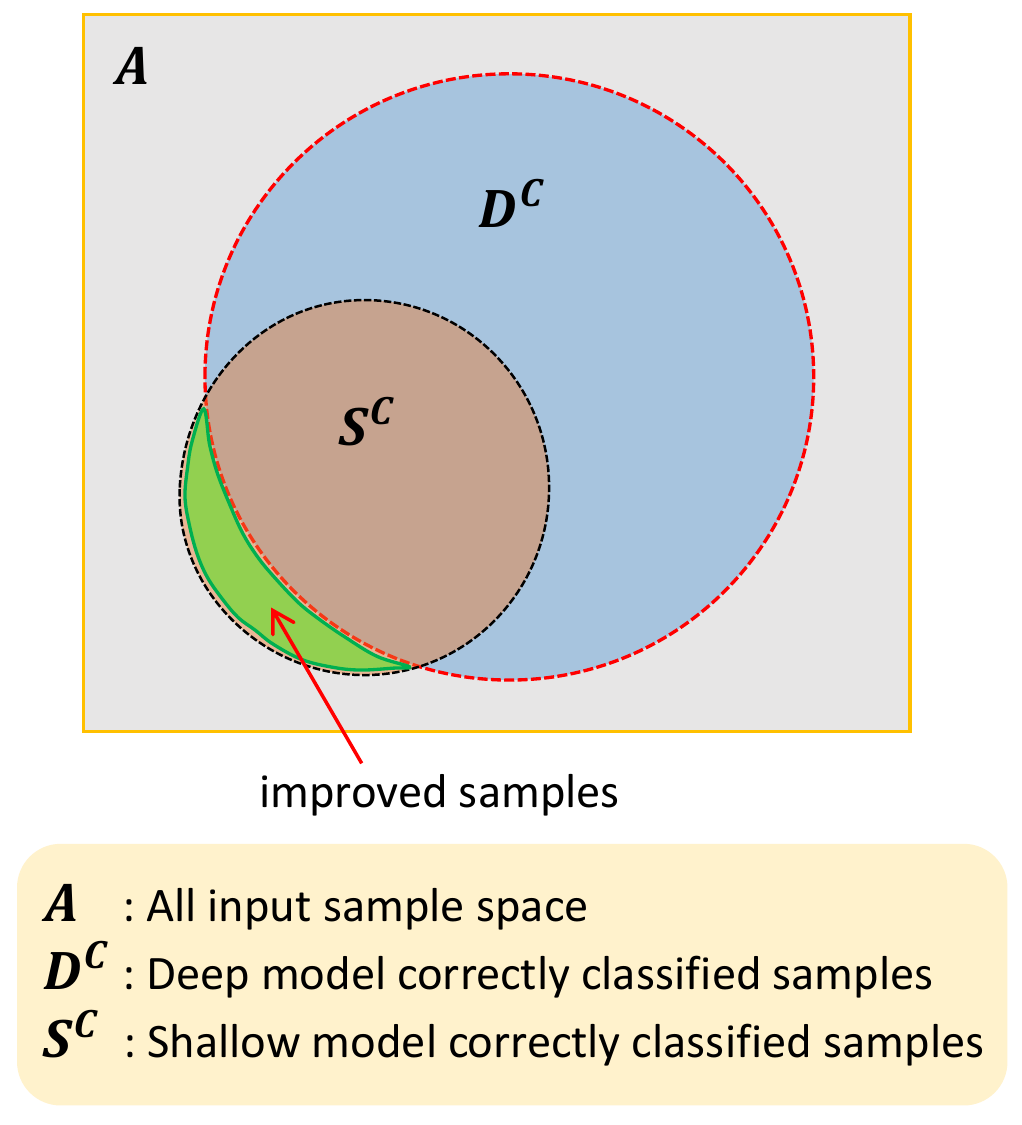}
  \caption{Diagram of the relationship of the samples in the dataset.}
  \label{fig5}
  \vspace{-1em}  % 調整上下文間距
\end{wrapfigure}

We can use CLCNet in a simple cascade structure system, in which we stack two (or several) classification models, as shown in Fig.\ref{fig1}. The model with fewer computations is called shallow model. The system will use it to classify first, and then input the classification result to CLCNet to predict whether it is classified correctly. If the confidence score output by CLCNet is higher than threshold, then we will directly accept the classification result and will not continue to the next steps, the threshold is a hyperparameter. The model with higher computation cost will be called deep model. When CLCNet's confidence to the shallow model's classification result is less than the threshold, the input sample will continue to be classified by the deep model, and the classification result will be also input to CLCNet for evaluation and give another confidence score. Finally, we compare the confidence scores of two models’ classification results, and accept the result with the higher confidence.

Obviously, the higher the threshold we set, the more samples will be further handled by the deep model for classification, and the average FLOPs per sample of the system will be higher, but the accuracy will also be improved. By adjusting the threshold value, we can customize the accuracy or average FLOPs of the system.

It should be noted that the upper limit of the accuracy of the system is \emph{not} the accuracy of the deep model. We use $A$ to represent the entire testing set, assuming that $S^{C}$ is a subset of samples classified correctly by the shallow model, where $S^{C}\subseteq A$. Likewise, suppose $D^{C}$ is the subset of samples correctly classified by the deep model, where $D^{C}\subseteq A$, we show their relationship in Fig.\ref{fig5}. It can be seen that although $S^{C}$ and $D^{C}$ mostly overlap, but $S^{C} \not\subseteq D^{C}$, this means that there are still a small number of samples that the shallow model classifies correctly, but the deep model classifies incorrectly, that is, the green area in Fig.\ref{fig5}. In an ideal situation, CLCNet can correctly find that these samples have been classified correctly in the shallow model, and will not continue to input to the deep model for classification, or CLCNet gives these correct shallow model classification results higher confidence scores than incorrect deep model classification results, thus preventing these samples from being misclassified by the system. Therefore, the accuracy upper limit of the system will be greater than the accuracy of the deep model.

\section{Experiments}
\label{experiments}

\subsection{Training and evaluation}
\label{4-1}
We first select many SOTA classification models that have been pre-trained on ImageNet-1k \cite{deng2009imagenet}, and then we combined these classification models in pairs and used them in the cascade structure system as shallow model and deep model for testing.

For the training step of CLCNet, in order to test the robustness of CLCNet, we only use EfficientNet-B4 \cite{tan2019efficientnet} to classify 50,000 images in the ImagNet-1k validation set, and use the classification results as the dataset of CLCNet. If EfficientNet-B4 classifies a image correctly, the label corresponding to the classification result is 1. Otherwise, the label is 0. So the size of the dataset is also 50,000, and each sample is a 1000-dimensional vector. For fairness, we use 5-fold cross-validation to train CLCNet and evaluate the cascade structure system. Different from regular cross-validation, at each fold, we divide the dataset into two parts of 40,000 samples and 10,000 samples, the training and validation sets of CLCNet are only provided by the part of 40,000 samples (80\% of the 40,000 samples are used for training and 20\% are used for validation and testing).

We use common and simple training methods, including using standard regression loss function (mean squared error) and Adam \cite{kingma2014adam} (learning rate of 0.002) as the optimizer, and train until convergence. When we have completed the training of CLCNet in these 40,000 samples, we will use this CLCNet in the cascade structure system and evaluate the classification accuracy of the system in 10,000 images (images of the ImageNet-1k validation set correspond to the part of the previously divided 10,000 samples). After repeating five folds, we can get the average accuracy of the cascade structure system on all 50,000 images of the ImageNet-1k validation set. Finally, we will use grid search on the threshold to get the performance of the cascaded structure system at the desired accuracy or computational cost for comparison.

\subsection{Comparison with models of the same structure of different sizes}
\begin{figure*}
  \centering
    \begin{subfigure}{0.45\textwidth}
      \centering   
      \includegraphics[width=1\linewidth]{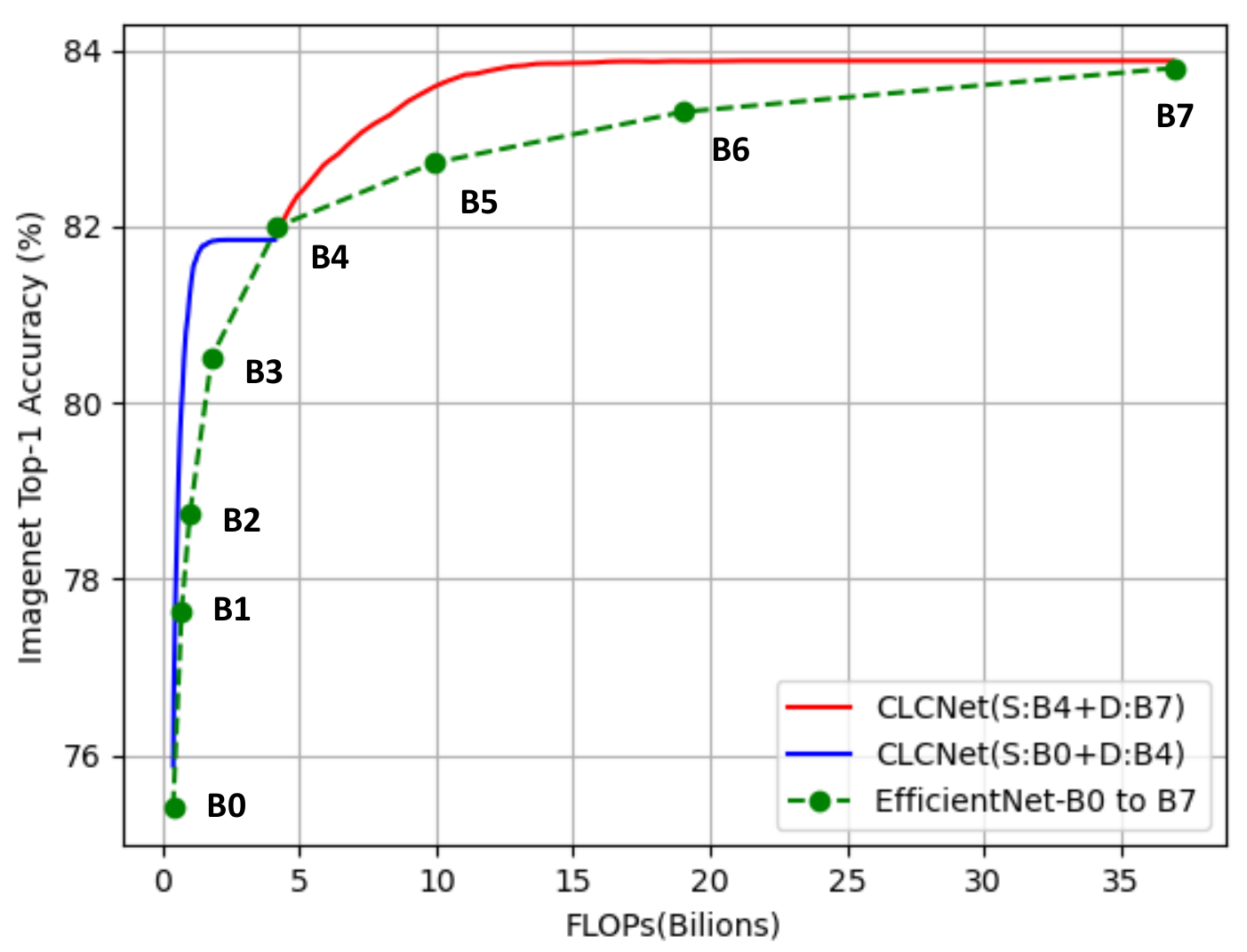}
        \caption{Comparison of CLCNet and EfficientNet.}
        \label{fig6-1}
    \end{subfigure}         %\hfill  % 这个\hfill指令为插入弹性长度的空白，看情况选择加不加。
    \hspace{1.5em}
    \begin{subfigure}{0.45\textwidth}
      \centering   
      \includegraphics[width=\linewidth]{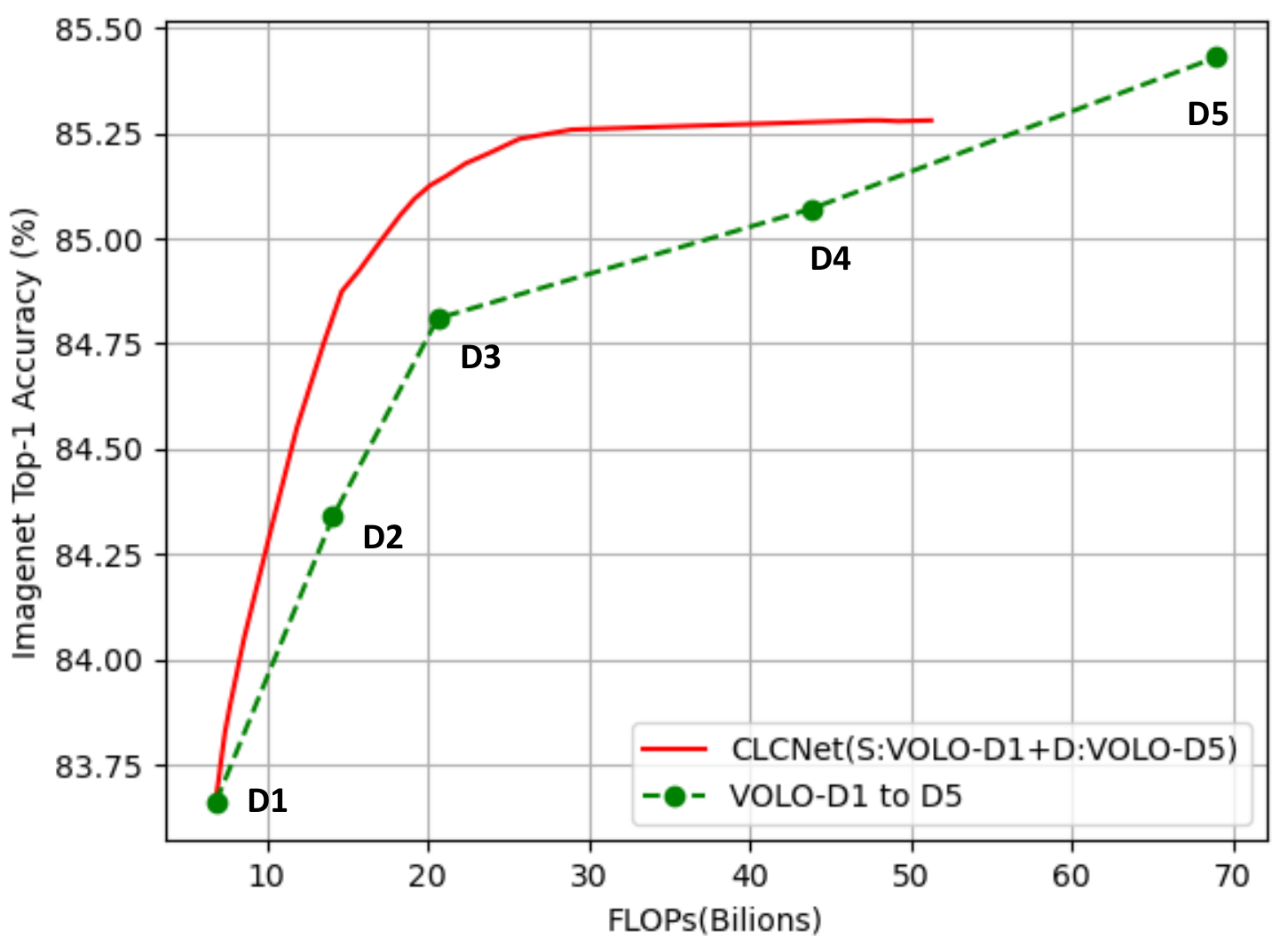}
        \caption{Comparison of CLCNet and VOLO.}
        \label{fig6-2}
    \end{subfigure}
\caption{Comparison of CLCNet and EfficientNet/VOLO on ImageNet-1K. The red and blue curves are obtained by setting FLOPs and accuracy as coordinates, and connecting those coordinates that CLCNet gets under different thresholds, S and D stand for shallow model and deep model.}
\label{fig6}
\end{figure*}

We first choose the classic EfficientNet \cite{tan2019efficientnet} series variants for testing. We use EfficientNet-B0/B4 as the shallow model of the cascade structure system and EfficientNet-B4/B7 as the deep model. For brevity, we will use CLCNet to denote the cascade structure system using CLCNet in following texts (If it represents the original CLCNet, it will be marked with (Fig.\ref{fig2}).). We set the accuracy or FLOPs of variants of different sizes of EfficientNet to the desired accuracy or computational cost, and then use grid search on the threshold to make the system just reach the requirement. In Tab.\ref{table1} we compare CLCNet and EfficientNet variants of different sizes under the same FLOPs or the same Top-1 accuracy on ImageNet-1k. Besides, we also draw the comparison in Fig.\ref{fig6-1}.

Obviously, for intermediate size variants of EfficientNet, we found that under the same FLOPs, CLCNet has higher accuracy, and under the same accuracy, CLCNet only needs lower FLOPs. Further, at the bottom of Tab.\ref{table1}, the highest accuracy of CLCNet even slightly exceeds the accuracy of the deep model B7, and only half of its average FLOPs is required, which is also consistent with Fig.\ref{fig5}.

\begin{table}
\centering
\caption{Comparison of CLCNet and EfficientNet on ImageNet-1k. Bold indicates the value of the metric is close to EfficientNet variant, S and D stand for shallow model and deep model respectively, and we use CLCNet to denote the cascade structure system using CLCNet.}
\vspace{1em}
\label{table1}
\setlength{\tabcolsep}{6mm}{
\begin{tabular}{@{}l||c|c|c@{}}
\toprule[1.5pt]
Model             & Top-1 Acc.       & Threshold & FLOPs per image \\ \midrule[1pt]
EfficientNet-B0   & 75.40\%          & \#\#      & 0.39B           \\ \midrule
EfficientNet-B1   & 77.64\%          & \#\#      & 0.70B           \\
CLCNet (S:B0+D:B4) & \textbf{77.74\%} & 0.19      & \textbf{0.74B}  \\ \midrule
EfficientNet-B2   & 78.73\%          & \#\#      & 1.0B            \\
CLCNet (S:B0+D:B4) & 79.06\%          & 0.27      & \textbf{0.996B} \\
CLCNet (S:B0+D:B4) & \textbf{78.71\%} & 0.25      & 0.933B          \\ \midrule
EfficientNet-B3   & 80.52\%          & \#\#      & 1.8B            \\
CLCNet (S:B0+D:B4) & 81.19\%          & 0.43      & \textbf{1.77B}  \\
CLCNet (S:B0+D:B4) & \textbf{80.50\%} & 0.39      & 1.42B           \\ \midrule
EfficientNet-B4   & 82.00\%          & \#\#      & 4.2B            \\
CLCNet (S:B4+D:B7) & \textbf{82.02\%} & 0.05      & \textbf{4.27B}  \\ \midrule
EfficientNet-B5   & 82.72\%          & \#\#      & 9.9B            \\
CLCNet (S:B4+D:B7) & 83.59\%          & 0.45      & \textbf{9.94B}  \\
CLCNet (S:B4+D:B7) & \textbf{82.75\%} & 0.27      & 6.1B            \\ \midrule
EfficientNet-B6   & 83.30\%          & \#\#      & 19B             \\
CLCNet (S:B4+D:B7) & 83.88\%          & 0.83      & \textbf{18.58B} \\
CLCNet (S:B4+D:B7) & \textbf{83.42\%} & 0.39      & 8.95B           \\ \midrule
EfficientNet-B7   & 83.80\%          & \#\#      & 37B             \\
CLCNet (S:B4+D:B7) & \textbf{83.88\%}          & 0.83      & 18.58B          \\ \bottomrule[1.5pt]
\end{tabular}}
\end{table}

Next, we want to test the robustness of CLCNet. We directly combine the CLCNet (Fig.\ref{fig2}) trained by the dataset generated by EfficientNet-B4 with the VOLO-D1/D5 \cite{yuan2021volo} \emph{without} retraining CLCNet (Fig.\ref{fig2}), and also compare the system with VOLO variants of different sizes. The results are listed in Tab.\ref{table2}, and Fig.\ref{fig6-2} also draws their comparison. The results show that although CLCNet does not exceed the accuracy of the deep model, it still achieves better performance compared to the intermediate size variants of VOLO.

\begin{table}
\centering
\caption{Comparison of CLCNet and VOLO on ImageNet-1k. Bold indicates the value of the metric is close to VOLO variant, S and D stand for shallow model and deep model respectively.}
\vspace{1em}
\label{table2}
\setlength{\tabcolsep}{6mm}{
\begin{tabular}{@{}l||c|c|c@{}}
\toprule[1.5pt]
Model              & Top-1 Acc.       & Threshold & FLOPs per image \\ \midrule[1pt]
VOLO-D1            & 83.66\%          & \#\#      & 6.8B            \\
CLCNet (S:D1+D:D5) & \textbf{83.69\%} & 0.15      & \textbf{6.89B}  \\ \midrule
VOLO-D2            & 84.34\%          & \#\#      & 14.1B           \\
CLCNet (S:D1+D:D5) & 84.75\%          & 0.47      & \textbf{13.53B} \\
CLCNet (S:D1+D:D5) & \textbf{84.31\%} & 0.39      & 10.28B          \\ \midrule
VOLO-D3            & 84.81\%          & \#\#      & 20.6B           \\
CLCNet (S:D1+D:D5) & 85.126\%         & 0.71      & \textbf{20.14B} \\
CLCNet (S:D1+D:D5) & \textbf{84.87\%} & 0.51      & 14.64B          \\ \midrule
VOLO-D4            & 85.01\%          & \#\#      & 43.8B           \\
CLCNet (S:D1+D:D5) & 85.27\%            & 0.94      & \textbf{41.21B} \\
CLCNet (S:D1+D:D5) & \textbf{85.06\%} & 0.63      & 18.30B          \\ \midrule
VOLO-D5            & 85.43\%          & \#\#      & 69.0B           \\
CLCNet (S:D1+D:D5) & \textbf{85.28\%} & 0.95      & 47.43B          \\ \bottomrule[1.5pt]
\end{tabular}}
\vspace{1em}
\end{table}

It should be noted that the FLOPs of our proposed CLCNet (Fig.\ref{fig2}) is \emph{2.7M}, which is almost negligible with the \emph{Bilion} level of the compared classification models.

\subsection{Comparison with general ensemble modeling}
In addition to using different size variants of a model, we can use models with completely different structures to put into CLCNet for testing, and we will compare with the general ensemble modeling. In order to simulate the real usage scenarios of general ensemble modeling, we try to select SOTA classification models with similar accuracy and large structural differences.

\paragraph{Retrain CLCNet with shallow model and deep model}
We also tried to use the current shallow model and deep model to retrain CLCNet (Fig.\ref{fig2}) for better performance. Specifically, we use the current shallow model and deep model to classify 50,000 images of the ImageNet validation set, respectively, and combine the classification results of the two models as a dataset of 100,000 samples. Then use the same cross-validation as in Sec.\ref{4-1}, that is, use four-fifths of the data to train and validate CLCNet (Fig.\ref{fig2}) each time (80,000 samples, 40,000 ImageNet validation set images are classified by two models), and get a CLCNet (Fig.\ref{fig2}) weight. Use this weight to evaluate the accuracy of the remaining one-fifth of the data on the cascade structure system (the remaining 10,000 ImageNet validation set images), and repeat five times to cover the entire ImageNet validation set. We use (retrain) as a marker in the Tab.\ref{table3} for distinction.

We compare the performance of single models, CLCNet and general ensemble modeling on ImageNet-1k in Tab.\ref{table3}. And we use classical ensemble averaging as a general ensemble modeling method for comparison, which means that two models infer an input sample at the same time, and add the outputs of the two models as the output result. For CLCNet, we used multiple pairwise model combinations. For example, we first used the base variant of Vision Transformer (ViT-B/16) \cite{dosovitskiy2020image} as the shallow model and the EfficientNet-B7 trained by Noisy Student \cite{xie2020self} as the deep model, because B7 has higher FLOPs. In addition to the above two models, we also used other model combinations and have been listed in Tab.\ref{table3}.

We found that CLCNet can achieve competitive performance to general ensemble modeling, surpassing the accuracy of the single models, but only using much less average FLOPs than general ensemble modeling.

\begin{table}
\centering
\caption{Comparison of CLCNet and general ensemble modeling (GEM) on ImageNet-1k. Bold indicates fewer computations or higher accuracy. Metrics for the same model may be inconsistent with other tables due to different training methods or input image sizes. (retrain) indicates that CLCNet (Fig.\ref{fig2}) is retrained with the current shallow model and deep model.}
\vspace{1em}
\label{table3}
\begin{tabular}{@{}l||c|c|c@{}}
\toprule[1.5pt]
Model                             & Top-1 Acc.       & Threshold & FLOPs per image \\ \midrule[1pt]
EfficientNet-B7 (Noisy Student) \cite{xie2020self}   & 85.60\%          & \#\#      & 37B             \\
ViT-B/16 \cite{dosovitskiy2020image}                          & 85.22\%          & \#\#      & 33.03B          \\
ConvNeXt-L \cite{liu2022convnet}                        & 85.04\%          & \#\#      & 34.4B           \\
VOLO-D3 \cite{yuan2021volo}                          & 85.71\%          & \#\#      & 67.9B           \\ \midrule
CLCNet (S:ViT+D:EffNet-B7)        & 86.42\%          & 0.51      & \textbf{40.75B} \\
CLCNet (S:ViT+D:EffNet-B7) (retrain)       & \textbf{86.61\%}          & 0.96     & 51.93B \\
ViT+EffNet-B7 (GEM)               & 86.55\% & \#\#      & 70.03B          \\ \midrule
CLCNet (S:ConvNeXt-L+D:EffNet-B7) & 86.39\%          & 0.83      & 46.12B \\
CLCNet (S:ConvNeXt-L+D:EffNet-B7) (retrain)& \textbf{86.42}\%          & 0.73      & \textbf{45.43B} \\
ConvNeXt-L+EffNet-B7 (GEM)        & \textbf{86.42\%} & \#\#      & 71.4B           \\ \midrule
CLCNet (S:ViT+D:VOLO-D3)          & 86.28\%          & 0.75      & \textbf{56.55B} \\
CLCNet (S:ViT+D:VOLO-D3) (retrain)         & \textbf{86.46\%}          & 0.85      & 57.46B \\
ViT+VOLO-D3 (GEM)                 & 86.44\% & \#\#      & 100.93B         \\ \midrule
CLCNet (S:ViT+D:ConvNeXt-L)       & 86.00\%         & 0.75      & \textbf{44.95B} \\
CLCNet (S:ViT+D:ConvNeXt-L) (retrain)      & 86.02\%         & 0.97      & 51.66B \\
ViT+ConvNeXt-L (GEM)               & \textbf{86.18\%} & \#\#      & 67.43B          \\ \bottomrule[1.5pt]
\end{tabular}
\vspace{0em}
\end{table}

\section{Conclusion}

In this paper, we propose a CLCNet that can predict confidence scores for the classification results in arbitrary dimension, and the CLCNet can be used in a simple cascade structure system, which is able to approximate or even exceed the performance of general ensemble modeling, while required much less computation than general ensemble modeling. And the models in the system are replaceable, new SOTA models can be replaced for better results. Further, by adjusting the threshold of the system, the average FLOPs of the system inference can be specified.

\bibliography{ref}
\bibliographystyle{unsrt}

%%%%%%%%%%%%%%%%%%%%%%%%%%%%%%%%%%%%%%%%%%%%%%%%%%%%%%%%%%%%

%%%%%%%%%%%%%%%%%%%%%%%%%%%%%%%%%%%%%%%%%%%%%%%%%%%%%%%%%%%%

\end{document}